%% file: main.tex
\documentclass[12pt,conference]{inatel} 
\input{configuracoes/includes.tex}
\input{configuracoes/config.tex}
\usepackage{float}
\begin{document}

  \title{Poaching Hotspot Identification Using \\Satellite Imagery}
  \author
  {
    \IEEEauthorblockN{Aryan Pandhi (ap663), Shrey Baid (sb857), Sanjali Jha (sj373)}
  }

\maketitle
\input{secoes/motivation}
\input{secoes/probdef}
\input{secoes/relatedwork}
\input{secoes/method}
\input{secoes/experimental}
\input{secoes/eval}

\printbibliography
\end{document}

%% file: configuracoes/includes.tex
\usepackage[%usado para determinar medidas
top=1.78cm,
bottom=1.78cm,
left=1.65cm,
right=1.65cm,
headsep=0cm,
]{geometry}
\usepackage{times}
\usepackage{enumitem}%redefinir espacos itemize
\usepackage{graphicx}
\usepackage{url,hyperref}
\usepackage[utf8]{inputenc}
\usepackage{float}%mais controle para manipular figuras
\usepackage{caption}%manipular legenda da figura e tabela
\usepackage{mathtools}%equacoes
\usepackage[hang,flushmargin]{footmisc} 
\usepackage{xcolor}
\usepackage{wrapfig} %usado para envolver figura com texto
\usepackage{fancyhdr}%criacao do cabecalho
\usepackage{etoolbox}
\usepackage[export]{adjustbox}%mais controle para ajustar tamanho da tabela
\usepackage{comment}%ambiente para comentario
\usepackage{relsize} %usado por comandos \mathlarger

\usepackage[
    style=numeric,
    sorting=none,
    maxbibnames=10]{biblatex}
\usepackage[english]{babel}

%% file: configuracoes/config.tex
\makeatletter
\newcommand{\linebreakand}{%
  \baselineskip 
  \end{@IEEEauthorhalign}
  \hfill\mbox{}\par
  \mbox{}\hfill\begin{@IEEEauthorhalign}
}
\makeatother

\addto\captionsbrazil{
  \renewcommand{\abstractname}{Abstract}
  \renewcommand{\tablename}{TABELA}
}

\addto\captionsenglish{
  \renewcommand{\tablename}{TABLE}
}

%% file: secoes/motivation.tex
\section{Motivation}

Elephant Poaching in African countries has been a decade-old problem. So much so that African Forest Elephants are now listed as an endangered species, and African Savannah Elephants as critically endangered by the IUCN (International Union for Conservation of Nature). \cite{iucn_2021} Elephants are hunted primarily for their ivory tusks which caused many elephants to be born tuskless as a genetic modification for survival. \cite{maron_2021} Data gathered by recent studies shows that though poaching methods remain the same, the poaching grounds are rather dynamic. Poachers have shifted to areas with less ranger patrols and several other factors like watering holes, seasons, altitude etc. cause constant shifts in poaching hotspot locations. \cite{phenthorne_2020} After a period of low poaching from 2000-2014, poaching numbers in African countries are now on the rise again - WWF (World Wildlife Foundation) says there are 20,000 elephants poached annually \cite{wwf}. In African countries, anti-poaching efforts are concentrated near towns, while a majority of poaching occurs in the deserted regions. All of these factors result in the need for a Computer Vision Model to identify poaching hotspots through locating the geographic indicators of favorable poaching regions. A CV model eliminates the need to manually track poachers and account for the environmental factors to deploy resources and its combination with satellite imagery allows us to survey large areas without disturbing local species or cross border aviation restrictions. 

%% file: secoes/probdef.tex
\section{Problem Definition}
The problem statement can be defined as the following: \textit{Generating a heatmap of elephant poaching probabilities by identifying proximity to geographic factors favorable to poaching activities, in Africa, using Deep Learning.}\\

The following factors are found to affect poaching activities \cite{ferreguetti} : 
\begin{itemize}
  \item[] $a_h$ : amount of human settlement (built up) nearby
  \item[] $a_t$ : amount of trees nearby
  \item[] $a_g$ : amount of grassland nearby
  \item[] $d_f$ : distance from forest
  \item[] $d_w$ : distance from herbaceous wetland
\end{itemize}

The input to the problem is a square plot of land that can be represented by the geographic coordinates of its four corners. The square plot of land can be split into an $n \times n$ grid $\boldsymbol{G}$ of pixels represented by its satellite image. Each cell in $G$ would represent which geographical factor is present in the same pixel of the image. The following 4 factors are considered: herbaceous wetland, human settlements (built up), grass, trees.

An $n' \times n'$ grid $\boldsymbol{V}$ [$n' < n$] is created such that $\boldsymbol{V}[i,j]$ holds the values of $(a_h,a_t,a_g,d_f,d_w)$.

A function $f(a_h,a_t,a_g,d_f,d_w)$ is determined that outputs the probability of poaching. $f$ is then run on each cell of $\boldsymbol{V}$ to generate the probabilistic heat-map $\boldsymbol{P}$ i.e. $\boldsymbol{P}[i,j] = f(\boldsymbol{V}[i,j])$ 

%% file: secoes/relatedwork.tex
\section{Related Work}
While there have been prior attempts to automate the mitigation of illegal poaching, those studies fall short of the scale and usability of a deep learning model like a Convolution Neural Network deployed on satellite imagery data. Ferreguetti et al. (2018) \cite{ferreguetti} predicted potential poaching hotspots in Brazil by applying occupancy modelling to produce a spatially explicit diagnostic of poaching in protected areas. This was an attempt at predicting some factors that influence both the detectability and intensity of poaching such as distances from water resources and geographical accessibility, but the need to set up expansive camera trap grids and human access required to do the same significantly limited the scalability of this method. Moreover, the study only utilized single-season, single-species occupancy models which severely limit the applicability of the models themselves. 

As mentioned above, identifying potential hotspots involves detecting the geographical indicators like water bodies, settlements, forests. CNN and NDWI (Normalized Water Difference Index) Models have been used for water bodies identification in high resolution satellite images. Wang et al. (2020) \cite{WATERID} introduced using DenseNets (Densely Connected CNNs) for the same and generated higher accuracies. Corbane et al. (2020) \cite{corbane} use a multi-model approach with CNNs across time zones to detect human settlements in satellite images to account for variations in human culture. Mohanty et al (2020) \cite{mohanty} utilize state-of-the-art Mask RCNN to categorize roads, trees, and crops from satellite imagery.
			
Satellite imagery has recently emerged as a viable option to monitor wildlife populations as well, since large areas can be covered in short intervals, international surveys can be conducted with ease and without disturbing the local species. There have been prior attempts to utilize deep learning on satellite imagery to detect and count animals in different landscapes. Stapleton et al. (2014) \cite{stapleton} used satellite imagery to keep track of arctic wildlife, and Duporge et al. (2020) \cite{duporge} pioneered the detection of animals in heterogeneous landscapes to monitor population fluctuations and identify causes of decline. Their Faster RCNN pretrained on the COCO dataset \cite{coco} and  fine-tuned on images from the Worldview-3 and Worldview-4 satellites had an F2 score of 0.78 in heterogeneous areas compared to the human label accuracy of 0.77. While the study was successful in identifying elephants in complex heterogeneous topographies, it did not contribute to the prevention of poaching of those animals. In conjunction with the satellite imagery, we address the primary limitations in such approaches using a deep learning model by taking into account all of the most crucial environmental factors affecting the intensity and detectability of poaching to create a better predictive model.

%% file: secoes/method.tex
\section{Method}
The region of Northern Botswana has the highest poaching numbers in Africa and so we chose this region to create a probabilistic heatmap of poaching. The square region with the following geographical coordinates is chosen as our plot of land (latitude, longitude): ($-18$, $23$), ($-18$, $24$), ($-19$, $23$), ($-19$, $24$). The plot of land contains $12,000 \times 12,000$ pixels, which is also the size of grid $\boldsymbol{G}$. The method to generate a heatmap for poaching hotspots can be split into 3 parts - (A) Identification of the location of poaching factors -  herbaceous wetland, grassland, human settlements, and trees (B) Feature Engineering of Labeled Data, (C) Generation of the probabilistic heatmap $\boldsymbol{P}$ for the region.

\noindent \begin{center}
A. Identification of Poaching Factors\\
\end{center}

We initially used deep learning models to recognize poaching factors in satellite images. We used Google’s TensorFlow and Python’s open-source neural network package Keras for the majority of our implementation for these deep learning models. Our design process included the implementation of a basic CNN architecture, as well as the utilization of pre-trained models such as VGG-16 \cite{simonyan_zisserman}, Inception V3 \cite{szegedy_vanhoucke_ioffe_shlens_wojna_2015}, and ResNet-50 \cite{ieee} . Each model is used with inputs of images to get output scores for the various predetermined geographical factors we consider to influence elephant poaching - human built-structures, water bodies and forests.\\

\noindent1. Baseline Model

Our first implementation was a baseline CNN for reference that consisted of two convolution layers with ReLU with max pooling and dropout layers followed by two more such layers and another pair of max pooling and dropout layers. We trained this model on a set of tagged satellite images from UC Merced Land Use Dataset \cite{yang_newsam_2010} but it required a lot more data to effectively train this model to get usable results, it was faster to incorporate and fine-tune the pre-trained models and they yielded better classifications.\\

\noindent2. Deep Learning Architecture Incorporating Pre-Trained Models

Following our implementation of the baseline CNN architecture, we needed pre-trained models that have historically performed well and to incorporate these models, we used an architecture like that in Fig. 1.

\begin{figure}[h]
    \centering
    \fbox{\includegraphics[width=\columnwidth]{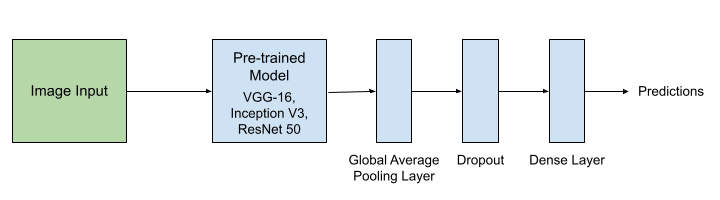}}
    \caption{Deep Learning Model Structure}
    \label{fig:incitel}
\end{figure}

For each pre-trained model, we add a global average pooling layer and dropout layer followed by a Dense layer to generate predictions. This architecture was inspired by TensorFlow’s Transfer Learning documentation, which allowed us to easily incorporate these pretrained models to our application. We trained these models both by freezing the weights on the pretrained models and training the added layers as well as by unfreezing the top 100 layers of the imported models and training those along with the added layers. To build these models, we used the following networks:\\

\noindent2.1. VGG-16

Originally developed at the University of Oxford, this network has 13 convolution layers and 3 densely connected layers that make up about 14 million parameters and achieved a 90.1$\%$ top-5 test accuracy on ImageNet. It has historically also proven effective at detecting objects like roads from remotely sensed data.\\

\noindent2.2. ResNet 50

ResNet 50 is a convolutional neural network developed at Microsoft that is 50 layers deep that make up about 23 million parameters. This model achieved a top-5 accuracy of 92.1$\%$ on ImageNet and has been previously applied to satellite and scene image classification.\\

\noindent2.3. Inception V3

Google’s Inception V3 network is 48 layers deep and focuses on being more computationally efficient than its predecessors. It can also be much deeper than VGG networks while still being more computationally efficient. \\

\noindent 3. Training and Evaluation 

We utilized a 70-20-10 train-test-validation split on a dataset procured from USGS National Map Urban Area Imagery collection and Landsat Satellite Images. We utilized transfer learning by importing models pretrained on the ImageNet dataset and fine tuning them on our dataset described above. While the ImageNet dataset consists of photographs and not satellite images, CNNs pretrained on that dataset have yielded effective results after fine-tuning them on other datasets. \cite{ieee} \cite{ganakwar_2020} \cite{shabbir} This process significantly shortened the training time. To evaluate, we used the F2 metric since it penalizes false negatives more so than false positives since we would like to identify all possible occurrences. Results are presented in Table 1. \\

\noindent 4. Factor Identification Sources

While transfer learning helped us in training our deep learning architecture by simply fine tuning the additional layers on top of the imported networks, we did not have sufficient computing power to train all of the deep learning models without freezing all, if not most of the layers of the imported models. To mitigate this, we chose the best performing frozen model, Inception V3, and trained it with several more unfrozen layers. This newly trained deep learning model achieves acceptable accuracy in detecting human built-up and grasslands. For the other environmental factors we considered i.e. distance from trees/forests, herbaceous wetlands and ubiquity of trees within the grid, we used WorldCover. 

WorldCover \cite{wc} is a landcover map of the world with a 10 meter resolution made from Sentinel-1 and 2 data. It splits each pixel into one of 11 land cover classes with a minimum overall accuracy of 75$\%$ \cite{worldcover}. The land cover classes important to us will be herbacious wetland and trees. Each 10 meter pixel of the chosen region in WorldCover will be a cell in the grid $\boldsymbol{G}$. We will let each cell $\boldsymbol{G}[i,j]$ be one of the 2 classes: trees, herbaceous wetland.

Shown below is an image of a plot of land in Botswana, the same plot of land on the WorldCover map, and a close up of a section of the WorldCover map with each pixel denoting a land form.

\begin{figure}[h]
    \centering
    \fbox{\includegraphics[width=0.8\columnwidth]{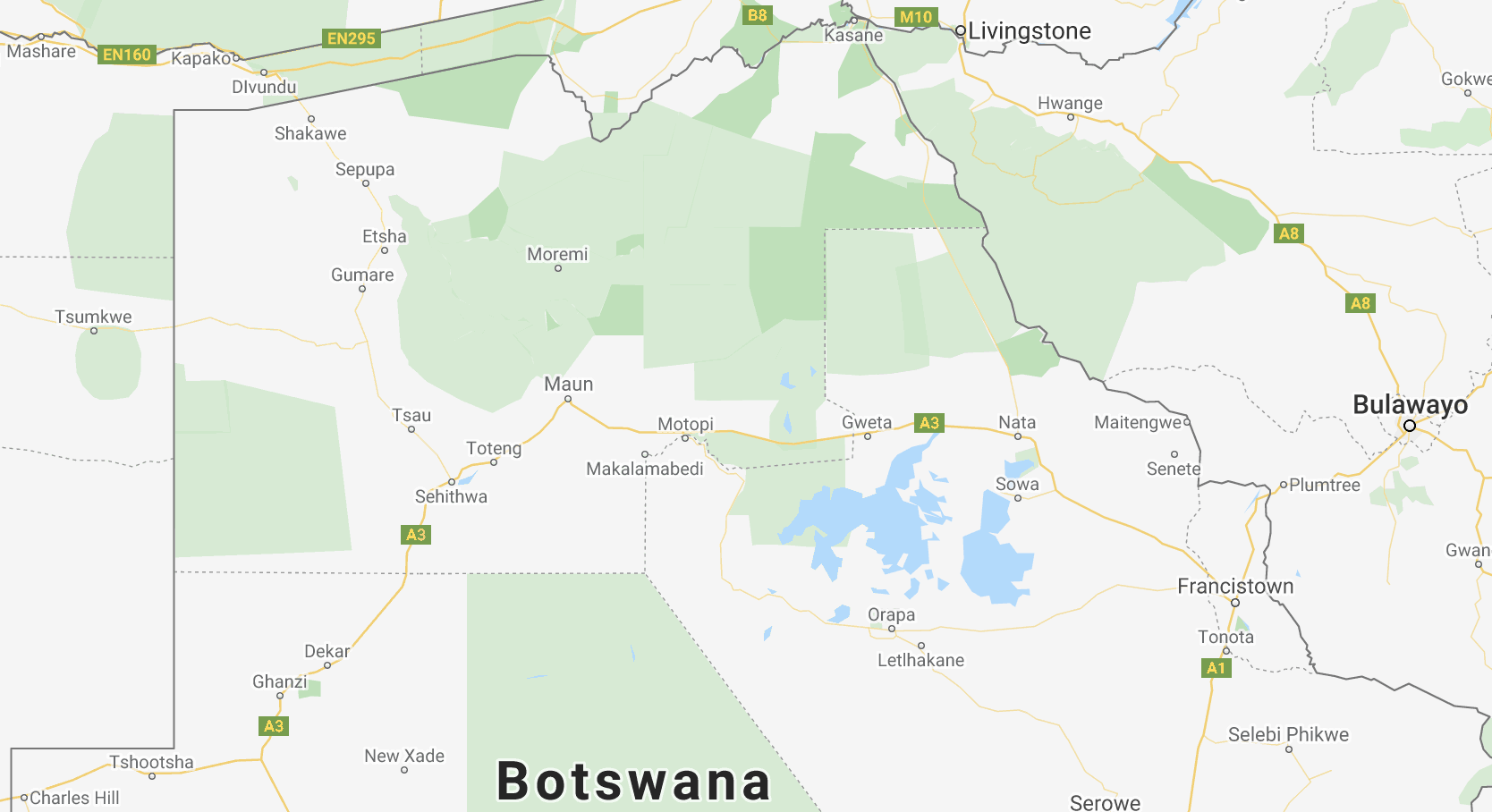}}
    \caption{Map of Northern Botswana Region}
    \label{fig:incitel}
\end{figure}

\begin{figure}[h]
    \centering
    \fbox{\includegraphics[width=0.8\columnwidth]{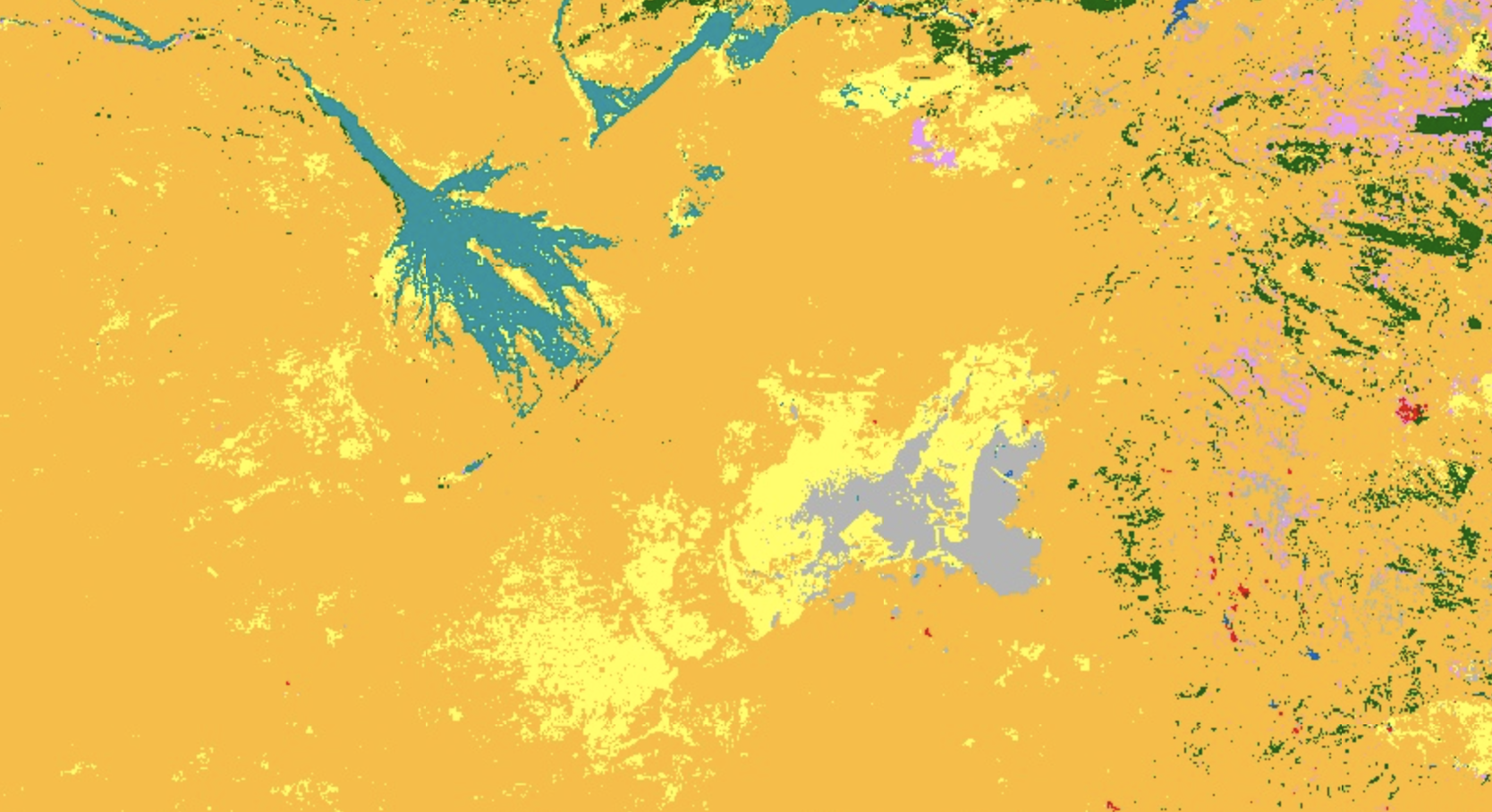}}
    \caption{Land Cover map of Northern Botswana Region}
    \label{fig:incitel}
\end{figure}

\begin{figure}[h]
    \centering
    \fbox{\includegraphics[width=0.8\columnwidth]{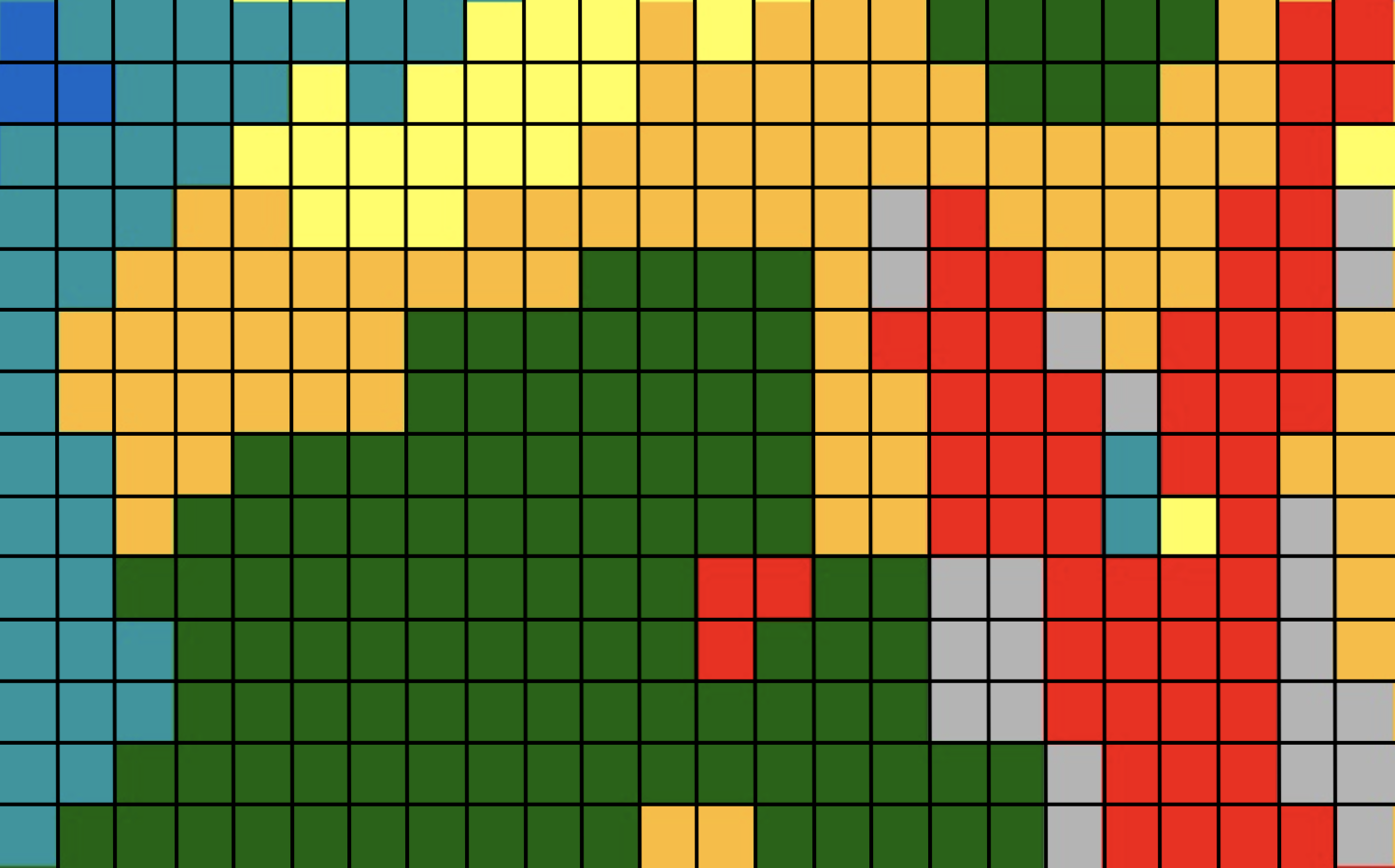}}
    \caption{A portion of a land cover grid of pixels}
    \label{fig:incitel}
\end{figure}

\newpage \noindent \begin{center}
B. Feature Engineering\\
\end{center}

Once we created the matrix $\boldsymbol{G}$, which represented a land use class for each pixel in the image, we created a feature matrix $\boldsymbol{V}$ using the following methodology: each $g \times g$ smaller window of pixels in $\boldsymbol{G}$ is grouped together to create one feature vector which includes the features $(a_h,a_t,a_g,d_f,d_w)$. The number of pixels in the $g \times g$ window containing human settlement, trees and grass are used to determine $a_h$, $a_t$, and $a_g$. Then for each cell in $\boldsymbol{V}$, the closest cell in $\boldsymbol{V}$ that contains trees or wetland is calculated using a grid search to determine $d_f$ and $d_w$. We chose $g=20$, which resulted in $V$ of size $600\times600$. Note that we set the distance to 0 if that factor is included in the current window. This method of creating $\boldsymbol{V}$ resulted in a smaller feature matrix, which allowed us to train our final classifiers more easily and led to better predictions over larger areas as opposed to a classifier predicting over each 10m pixel. 

\begin{figure}[h]
    \centering
    \fbox{\includegraphics[width=0.8\columnwidth]{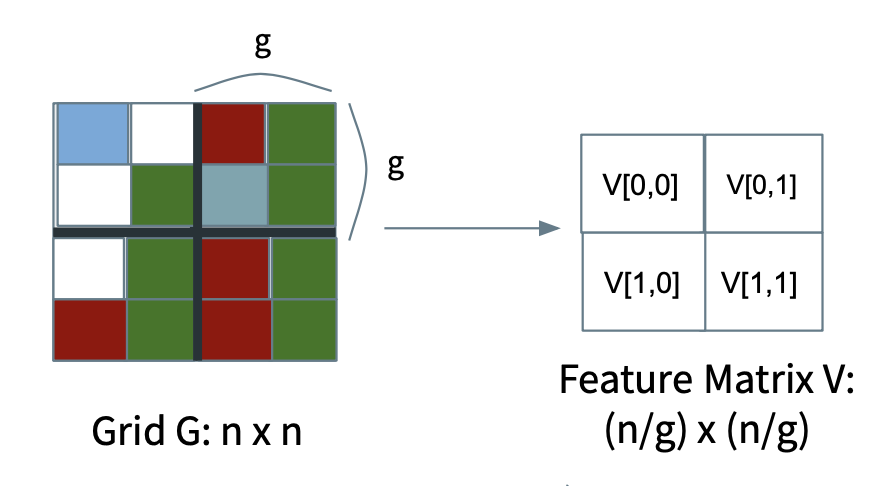}}
    \caption{Feature Matrix Generation}
    \label{fig:incitel}
\end{figure}

Now that we have the feature matrix, using data from Schlossberg et. al. \cite{evidenceof}, we constructed labels based on locations of previously verified poached elephant carcasses. We had 40 locations of past poaching incidents in our chosen plot of land and used the feature vectors in $\boldsymbol{V}$ for those particular locations as $+1$ labeled data. For each cell $c$ in $\boldsymbol{V}$, the distance to the nearest poaching incident cell determined if the feature vector of $c$ was to be used as training data or not. For each cell $c$ in $\boldsymbol{V}$, if a poaching incident cell was adjacent to it, the feature vector of $c$ was used as $+0.9$ labeled data. Similarly, if a poaching incident cell was two cells away, the feature vector was used as $+0.8$ and so on. The cells in $\boldsymbol{V}$ that were greater than 440 cells away from a poaching incident cell were used as $0$ labeled data. Therefore, for any cell $c$, if distance to incident was $<= 10$ cells, it was used as positive labeled data and if distance was $> 400$ cells, it was used as zero labeled data. with a $600\times600$ feature matrix $\boldsymbol{V}$ and 40 past poaching incidents, we were able to create $6606$ positive label training points (label $>0$) and $7350$ zero label training points (label $=0$).

\noindent \begin{center}
C. Training Models\\
\end{center}

Now that we have training data, it was split into $60-20-20$ training, validation, test sets. We set it up as a regression problem, and the following ML models were trained on the data: (i) Random Forest Regressor, (ii) Support Vector Regressor, and (iii) Multi-layer Perceptron Regressor.

%% file: secoes/experimental.tex
\section{Results}

% \begin{table}[h]
% \centering
% \caption{Results}
% \label{tab:tabela1}
% \begin{adjustbox}{max width=\columnwidth}
% \begin{tabular}{|c|c|c|c|}
% \hline
% Model$|$F2 Validation Score & River/Water Body & Forests & Human Built-up \\ \hline\hline
% Baseline CNN & 0.503 & 0.553 & 0.571\\
% VGG-16 & 0.603 & 0.611 & 0.638\\
% ResNet 50 & 0.621 & 0.657 & 0.661\\
% Inception V3 & 0.681 & 0.684 & 0.689\\\hline
% \end{tabular}
% \end{adjustbox}
% \end{table}
The critical step in making these predictions was feature engineering. As discussed above, we used the best performing frozen pre-trained deep learning architecture and further trained it to detect human built-up and grasslands. To detect wetlands and forests/trees, we used WorldCover. Results are presented in Table I.

\begin{table}[h]
\centering
\caption{Feature Engineering Results}
\label{tab:tabela1}
\begin{adjustbox}{max width=\columnwidth}
\begin{tabular}{|c|c|c|c|c|}
\hline
Feature & Model/Source Used & Test Acc.\\ \hline \hline
Human Built-up & Inception V-3 Architecture & 0.731 \\
Grasslands & Inception V-3 Architecture & 0.711 \\
Wetlands & WorldCover & 0.75 \\
Trees & WorldCover & 0.75 \\\hline
\end{tabular}
\end{adjustbox}
\end{table}

It became clear that our deep learning architecture performed best at detecting human built-up, an important factor that we saw also affects the predictions the most. 

We conducted grid search on the parameters of all three regressor models to improve our prediction accuracy and evaluated them using the validation set. The chosen configuration is summarized in Table II below.

\begin{table}[h]
\centering
\caption{Parameter Configuration}
\label{tab:tabela1}
\begin{adjustbox}{max width=\columnwidth}
\begin{tabular}{|c|c|c|}
\hline
Model & Parameter 1 & Parameter 2\\ \hline\hline
Random Forest Regressor & Max Depth = 6 & No. of Trees = 500\\
Support Vector Regressor & Kernel = rbf & $C$ = 2, $\epsilon$ = 0.01\\
Multi-layer Perceptron Regressor & Solver = lbfgs & Hidden Layers = [5, 5, 10, 3]\\\hline
\end{tabular}
\end{adjustbox}
\end{table}

The training and test accuracy of all three trained regressors are summarized in Table III below.

\begin{table}[h]
\centering
\caption{Regressor Results}
\label{tab:tabela1}
\begin{adjustbox}{max width=\columnwidth}
\begin{tabular}{|c|c|c|}
\hline
Model & Train $R^2$ & Test $R^2$\\ \hline\hline
Random Forest Regressor & 0.75 & 0.71\\
Support Vector Regressor & 0.42 & 0.41\\
Multi-layer Perceptron Regressor & 0.47 & 0.46\\\hline
\end{tabular}
\end{adjustbox}
\end{table}

Random Forest Regressor clearly outperforms the other two models with a test $R^2$ value of $0.71$. We believe that the Random Forest Regressors worked considerable better than the SVR because of the number of outlier points we had. The Multi-layer Perceptron Regressor (NN) did not perform as we expected, potentially due to the small number of poaching hotspot data points to train on. 

In order to understand the geographic factors that affected our model the most, we measured the importance of each feature using the Random Forest Model. We used permutation feature importance to measure the increase in the prediction error of the model after permuting each feature’s value. This was chosen over the alternative of calculating feature importance based on mean decrease in impurity due to the latter being misleading for high cardinality features. The results are shown in figure 6.

\begin{figure}[h]
    \centering
    \fbox{\includegraphics[width=0.8\columnwidth]{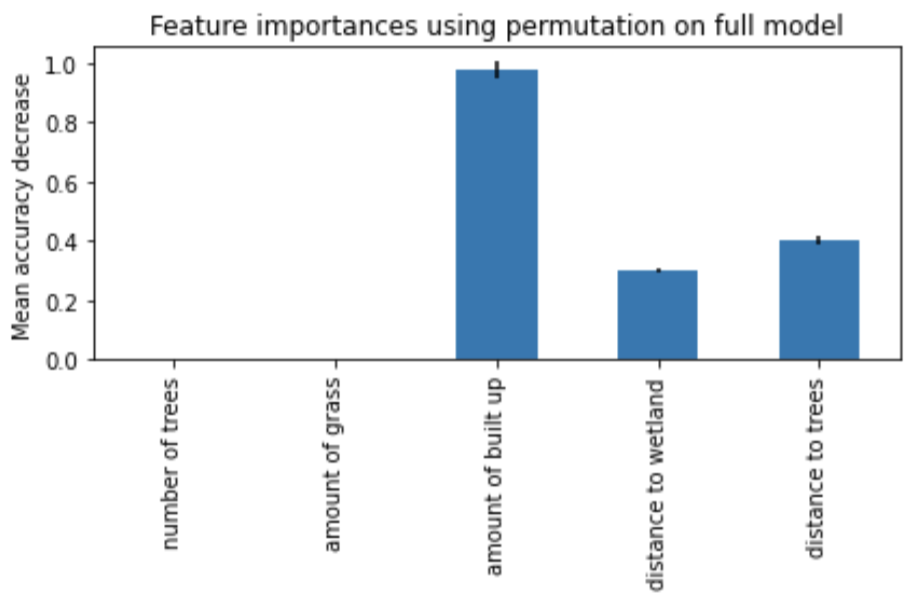}}
    \caption{Feature Importance}
    \label{fig:incitel}
\end{figure}

 As shown in figure 6, the most important feature according to the Random Forest Regressor is $a_b$ the amount of human settlement (built up) in the area. This is consistent with our  predictions as poaching activities are conducted as far away from human settlements as possible, in order to avoid detection. The 2 features that had negligible importance were $a_t$ (amount of trees) and $a_g$ (amount of grass). We believe this was because 90\% of Northern Botswana is covered in savanna, so the amount of grass and trees would be consistent for each window in grid $\boldsymbol{G}$, and these features may be more relevant for other regions in Africa.

The heatmap generated for our chosen plot of land using the Random Forest Regressor is shown in Figure 7 below with dark regions representing higher probabilities.

\begin{figure}[h]
    \centering
    \fbox{\includegraphics[width=0.8\columnwidth]{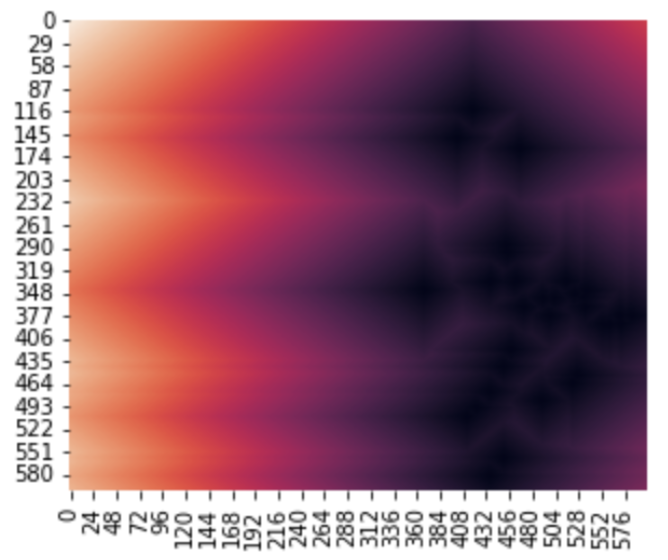}}
    \caption{Poaching Probability Heatmap}
    \label{fig:incitel}
\end{figure}

%% file: secoes/eval.tex
\section{Evaluation and Further Study}
The test accuracy of 0.71, when using the Random Forest Regressor, shows that it is possible to utilize Computer Vision with satellite imagery to correctly identify poaching hotspots. However, we can further improve the accuracy of predictions in multiple ways.
\begin{enumerate}
  \item A major restriction in our experiments was the lack of computing resources. With better computing resources, we could increase our training data to more countries in Africa and improve the performance of our feature extraction models. We would also be able to train more layers of the VGG-16, Inception V3, and ResNet 50 models gaining better accuracies and indirectly improving our hotspot predictions.
  \item Another limitation we faced was the availability of poaching data. A bulk of the poaching location data was not publicly available for us to utilize. The use of poaching hotspot locations from multiple years to generate more points for our training data would help improve the performance of the chosen regressors. 
  \item We used 5 geographical features to make poaching hotspot predictions. We could expand our features to other factors that affect poaching like temperature distribution, seasonal weather conditions, rainfall quantity, altitude for more accurate predictions. We could also include other features like number of elephant sightings and location of elephant carcasses as these are often correlated to poaching grounds. Further, we could assign weights to features based on the feature importance tables we generated using permutation.
\end{enumerate}
Nonetheless, the experiments we ran were a proof of concept to apply CV techniques to problems like poaching. This use case can be expanded to include all of Africa and help rangers predict poaching grounds to patrol and reduce the killings of elephants. \\
In addition to elephants, many other animals are constantly illegally poached all around the world. The poaching grounds for such animals too depend on different factors unique to each animal, and our model can be modified to predict poaching grounds for other wildlife animals as well. Other activities that are dependent on geographic features like illegal fishing, deforestation etc. can also be predicted and potentially reduced using similar concepts.

%% file: bibliography.bib
@article{WATERID,
author = {Wang, Guojie and Wu, Mengjuan and Wei, Xikun and Song, Huihui},
year = {2020},
month = {03},
pages = {795},
title = {Water Identification from High-Resolution Remote Sensing Images Based on Multidimensional Densely Connected Convolutional Neural Networks},
volume = {12},
journal = {Remote Sensing},
doi = {10.3390/rs12050795}
}

@misc{iucn_2021, title={African elephant species now Endangered and Critically Endangered - IUCN Red List}, url={https://www.iucn.org/news/species/202103/african-elephant-species-now-endangered-and-critically-endangered-iucn-red-list}, journal={IUCN}, year={2021}, month={Aug}}

@misc{maron_2021, title={Under poaching pressure, elephants are evolving to lose their tusks}, url={https://www.nationalgeographic.com/animals/article/wildlife-watch-news-tuskless-elephants-behavior-change}, journal={Animals}, publisher={National Geographic}, author={Maron, Dina Fine}, year={2021}, month={May}}

@misc{phenthorne_2020, title={Elephant Poaching in South Africa}, url={https://usfblogs.usfca.edu/sustainability/2020/05/15/elephant-poaching-in-south-africa/}, journal={Office of Sustainability - Student Blog}, author={Phenthorne}, year={2020}, month={May}}

@misc{wwf, title={Elephant}, url={https://www.worldwildlife.org/species/elephant}, journal={WWF}, publisher={World Wildlife Fund}}

@article{ferreguetti, title={One step ahead to predict potential poaching hotspots: Modeling occupancy and detectability of poachers in a neotropical rainforest}, volume={227}, DOI={10.1016/j.biocon.2018.09.009}, journal={Biological Conservation}, author={Ferreguetti, Átilla Colombo and Pereira-Ribeiro, Juliane and Prevedello, Jayme Augusto and Tomás, Walfrido Moraes and Rocha, Carlos Frederico Duarte and Bergallo, Helena Godoy}, year={2018}, pages={133–140}}

@article{corbane, title={Convolutional neural networks for global human settlements mapping from Sentinel-2 satellite imagery}, volume={33}, DOI={10.1007/s00521-020-05449-7}, number={12}, journal={Neural Computing and Applications}, author={Corbane, Christina and Syrris, Vasileios and Sabo, Filip and Politis, Panagiotis and Melchiorri, Michele and Pesaresi, Martino and Soille, Pierre and Kemper, Thomas}, year={2020}, pages={6697–6720}}

@article{coco, title={Microsoft COCO: Common Objects in Context}, DOI={10.1007/978-3-319-10602-1_48}, journal={Computer Vision – ECCV 2014 Lecture Notes in Computer Science}, author={Lin, Tsung-Yi and Maire, Michael and Belongie, Serge and Hays, James and Perona, Pietro and Ramanan, Deva and Dollár, Piotr and Zitnick, C. Lawrence}, year={2014}, pages={740–755}}

@article{mohanty, title={Deep Learning for Understanding Satellite Imagery: An Experimental Survey}, volume={3}, DOI={10.3389/frai.2020.534696}, journal={Frontiers in Artificial Intelligence}, author={Mohanty, Sharada Prasanna and Czakon, Jakub and Kaczmarek, Kamil A. and Pyskir, Andrzej and Tarasiewicz, Piotr and Kunwar, Saket and Rohrbach, Janick and Luo, Dave and Prasad, Manjunath and Fleer, Sascha and et al.}, year={2020}}

@article{duporge , title={Using very high-resolution satellite imagery and deep learning to detect and count African elephants in heterogeneous landscapes}, DOI={10.1101/2020.09.09.289231}, journal={ZSL Publication}, author={Duporge, Isla and Isupova, Olga and Reece, Steven and Macdonald, David W. and Wang, Tiejun}, year={2020}}

@article{stapleton, title={Polar Bears from Space: Assessing Satellite Imagery as a Tool to Track Arctic Wildlife}, volume={9}, DOI={10.1371/journal.pone.0101513}, number={7}, journal={PLoS ONE}, author={Stapleton, Seth and Larue, Michelle and Lecomte, Nicolas and Atkinson, Stephen and Garshelis, David and Porter, Claire and Atwood, Todd}, year={2014}}

@misc{simonyan_zisserman, title={Very Deep Convolutional Networks for Large-Scale Image Recognition}, url={https://arxiv.org/abs/1409.1556}, journal={arXiv.org}, author={Simonyan, Karen and Zisserman, Andrew}, year={2015}, month={Apr}}

@misc{szegedy_vanhoucke_ioffe_shlens_wojna_2015, title={Rethinking the Inception Architecture for Computer Vision}, url={https://arxiv.org/abs/1512.00567}, journal={arXiv.org}, author={Szegedy, Christian and Vanhoucke, Vincent and Ioffe, Sergey and Shlens, Jonathon and Wojna, Zbigniew}, year={2015}, month={Dec}}

@misc{ieee, url={https://ieeexplore.ieee.org/document/7780459/citations}, journal={IEEE Xplore Temporarily Unavailable}}

@article{yang_newsam_2010, title={Bag-of-visual-words and spatial extensions for land-use classification}, DOI={10.1145/1869790.1869829}, journal={Proceedings of the 18th SIGSPATIAL International Conference on Advances in Geographic Information Systems - GIS 10}, author={Yang, Yi and Newsam, Shawn}, year={2010}}

@article{ganakwar_2020, title={Convolutional Neural Network-VGG16 for Road Extraction from Remotely Sensed Images}, volume={8}, DOI={10.22214/ijraset.2020.30796}, number={8}, journal={International Journal for Research in Applied Science and Engineering Technology}, author={Ganakwar, Prajakta}, year={2020}, pages={916–922}}

@article{shabbir, title={Satellite and Scene Image Classification Based on Transfer Learning and Fine Tuning of ResNet50}, volume={2021}, DOI={10.1155/2021/5843816}, journal={Mathematical Problems in Engineering}, author={Shabbir, Amsa and Ali, Nouman and Ahmed, Jameel and Zafar, Bushra and Rasheed, Aqsa and Sajid, Muhammad and Ahmed, Afzal and Dar, Saadat Hanif}, year={2021}, pages={1–18}}

@misc{worldcover, title={Objectives}, url={https://esa-worldcover.org/en/about/objectives}, journal={WORLDCOVER}}

@misc{wc, title={WorldCover}, url={https://esa-worldcover.org/en}, journal={WORLDCOVER}}

@article{evidenceof,
author = {Schlossberg, Scott and Chase, Michael and Sutcliffe, Robert},
year = {2019},
month = {06},
pages = {},
title = {Evidence of a Growing Elephant Poaching Problem in Botswana},
volume = {29},
journal = {Current Biology},
doi = {10.1016/j.cub.2019.05.061}
}
